\documentclass[11pt,a4paper]{article}
\usepackage[final]{conf_paper/naacl2021}
\usepackage{times}
\usepackage{latexsym}
\usepackage{epsfig}
\usepackage{graphicx}
\usepackage{pgfplots}
\usepackage{amsmath}
\usepackage{amssymb}
\usepackage{graphicx}
\usepackage{subcaption}
\usepackage{booktabs}
\usepackage{soul}
\usepackage{multirow}
\usepackage{threeparttable}
\usepackage[cellcolor]{colortbl}
\usepackage{longtable}
\usepackage{array}
\usepackage{bbm}
\usepackage[toc,page]{appendix}
\usepackage[ruled,vlined]{algorithm2e}
\usepackage{algpseudocode}
\pgfplotsset{compat=1.16} 
\usepackage{lipsum} 

\usepackage{microtype}

\title{Supervised Learning in the Presence of Noise: Application in ICD-10 Code Classification}

\author{
Youngwoo Kim\textsuperscript{1} \\ University of Massachusetts Amherst 
\And
Cheng Li\textsuperscript{2} \\ CodaMetrix
\AND
 Bingyang Ye\textsuperscript{2}  \\ CodaMetrix  
  \And
  Amir Tahmasebi\textsuperscript{3}  \\ CodaMetrix \\ Enlitic 
  \And 
  Javed Aslam\textsuperscript{2} \\ CodaMetrix    
  \AND
   \normalfont\textsuperscript{1} \texttt{youngwookim@cs.umass.edu} \\
   \textsuperscript{2} \texttt{\{cheng,bingyang,jay\}@codametrix.com} \\ 
   \textsuperscript{3} \texttt{atahmasebi@enlitic.com}
 }

\date{}

\begin{document}
\maketitle
\begin{abstract}
ICD coding is the international standard for capturing and reporting health conditions and diagnosis for revenue cycle management in healthcare. Manually assigning ICD codes is prone to human error due to the large code vocabulary and the similarities between codes. Since machine learning based approaches require ground truth training data, the inconsistency among human coders is manifested as noise in labeling, which makes the training and evaluation of ICD classifiers difficult in presence of such noise. This paper investigates the characteristics of such noise in manually-assigned ICD-10 codes and furthermore, proposes a method to train robust ICD-10 classifiers in the presence of labeling noise. Our research concluded that the nature of such noise is systematic. Most of the existing methods for handling label noise assume that the noise is completely random and independent of features or labels, which is not the case for ICD data. Therefore, we develop a new method for training robust classifiers in the presence of systematic noise. We first identify ICD-10 codes that human coders tend to misuse or confuse, based on the codes' locations in the ICD-10 hierarchy, the types of the codes, and baseline classifier's prediction behaviors; we then develop a novel training strategy that accounts for such noise. We compared our method with the baseline that does not handle label noise and the baseline methods that assume random noise, and demonstrated that our proposed method outperforms all baselines when evaluated on expert validated labels.

\end{abstract}

\section{Introduction}
\begin{figure*}[ht]
    \centering
    \includegraphics[width=0.9\textwidth]{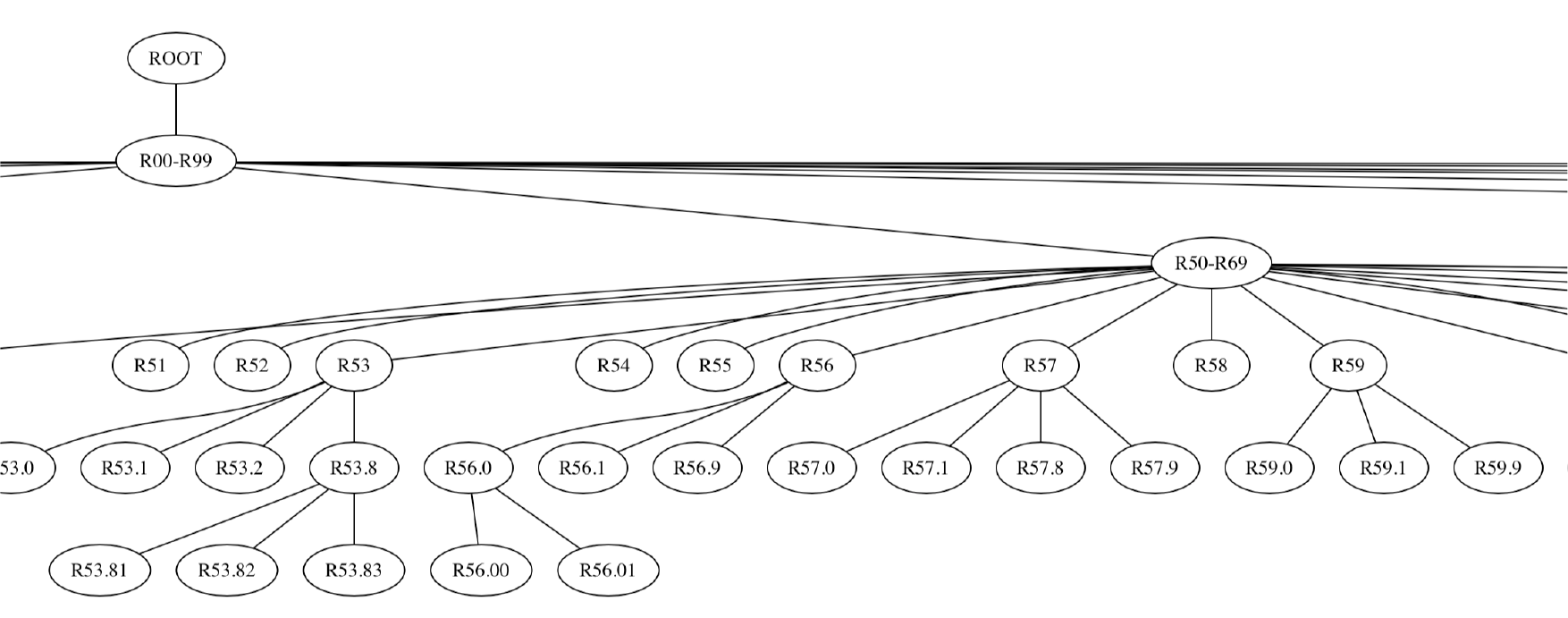}
    \caption{A snapshot of the ICD-10 hierarchy.}
    \label{fig: ICD-10 Hierarchy}
\end{figure*}

International Classification of Diseases (ICD) is a standard classification system for capturing and reporting diagnosis and medical conditions in clinical records for revenue cycle management in healthcare. It includes codes for diagnoses, symptoms, and procedures. The latest version of ICD, ICD-10, has over 70,000 codes. ICD-10 coding system consists of alphanumeric values with up to seven characters (e.g., R91.8), which are structured in a tree-like hierarchy \ref{fig: ICD-10 Hierarchy}. Addition of each character moves a code from one level to another in the hierarchy.

In Radiology, depending on the content, one or multiple ICD codes may be assigned to a single clinical record. Manually assigning ICD codes to health records is time-consuming due to large volume of health records. Furthermore, manual code assignment is a challenging task due to the large code vocabulary and the similarities between codes, which results in inconsistencies among human coders. Previous studies have shown that the inter-assessor agreement rate between human coders is merely 50\%~\cite{american2003icd, horsky2017accuracy, stausberg2008reliability}.

Recently, Computer-assisted Coding (CAC) systems have gained popularity as they utilize machine learning for ICD code prediction. Nevertheless, the inconsistency in human-driven labeling poses challenges in training and evaluation of ICD classifiers since the ground truth for training and evaluation may not be reliable.

This paper investigates and categorizes the noise in manually-assigned ICD codes, studies the effect of noisy data in the model training and evaluation, and finally, proposes a method to train robust ICD classifiers in the presence of labeling noise.  

In order to study the noise in ICD-10 coding, we used a relatively large dataset of radiology notes with corresponding ICD-10 codes from multiple clinical sites. We discovered systematic noise by analyzing labels assigned by two different human coders. One most prevalent noise pattern is over-using the codes defined for generic symptoms where the codes defined for specific diagnoses should have been used. Since rebuilding training data by expert coders to avoid such errors in labeling is expensive, it is desired to account for such noise in the ICD-10 classifier training that is capable of learning through noisy labels. Most of the existing methods for handling label noise assume that the noise is completely random and is independent of features or labels, which is not the case in ICD-10 data. 

In this work, we propose a new method for training robust classifiers in the presence of systematic noise. We first identify ICD-10 codes that human coders tend to misuse or confuse, based on the codes' descriptions, the types of the codes, and baseline classifier's prediction behaviors; we then develop a new training strategy that accounts for such noise. We compared our method with the baseline that does not handle label noise and the baseline methods that assume random and independent noise, and demonstrated that our proposed method outperforms all baselines when evaluated on expert validated labels.

Our contributions are as follows: 
\begin{enumerate}
    \item We identify systematic noise in ICD-10 codes annotated by human coders, and measure its negative impact on machine learning model's performance.
    \item We propose a method for detecting systematic noise automatically.
    \item We propose a robust training method to train classifiers on noisy labels and gain improvements when evaluated on validated labels.
\end{enumerate}

\section{Analyzing ICD-10 Coding Noise}
\label{sec:analyzing}
\begin{table*}[htbp]
\centering
\begin{tabular}{c|c|c|c}
\toprule
\textbf{Category} & \textbf{Condition} & \textbf{Ratio} & \textbf{Example ($S_o \rightarrow S_v)$}  \\ \hline
match & $S_o=S_v$ & 65.2\% & \{R91.8, R05\} $\rightarrow$ \{R91.8, R05\} \\
replacement  & $|S_o - S_v|=|S_v - S_o|=1$& 14.0\% & \{R05, J98.11\} $\rightarrow$ \{R05, J44.9\} \\ 
missing  & $S_o\subset S_v$ & 8.3\% & \{Z46.82\} $\rightarrow$ \{Z46.59, Z48.03 \} \\
extra    & $S_o \supset S_v$ & 7.8\% & \{M54.5, M41.86\} $\rightarrow$ \{M41.86\} \\
other       &  other & 4.6\% & \{M54.5\} $\rightarrow$ \{M47.816, M48.56XA\} \\ \bottomrule
 
\end{tabular}
\caption{Code-set level disagreement categories. $S_o$ is the original code set and $S_v$ is the validated code set. }
\label{tab:code_set_category}
\end{table*}
In order to handle the labeling noise during classifier training, we first analyze the nature of noise by collecting data with both the original noisy labels and validated/corrected labels by expert coders.

\subsection{Validating annotation}
We took a large collection of Radiology notes from multiple clinical sites annotated with ICD-10 codes (the latest version) by several human coders. We randomly sampled notes using stratified sampling to make sure different code sets are represented in the sample, and then asked expert coders to validate existing code annotations and make corrections if necessary. Each radiology note is reviewed by at least two expert coders. If the two expert coders disagree, a third expert coder who is more experienced is asked to make the final judgment. A total of 29,826 Radiology notes were validated through the schema described above. For noisy pattern analysis (section~\ref{sec:analyzing}), all the validated notes are used. For the model training and evaluations (section~\ref{sec:howdoeslabelnoise}), we selected one of the clinical sites and used only the notes from the selected site, which has 1,688 notes. 

We call the resulting code labels \textbf{validated labels} in contrast to the \textbf{original labels}. Accordingly the corresponding coders would be referred to as \textbf{validating coders} and \textbf{original coders}. In this research, we refer to the original labels as noisy labels and validated labels as clean labels even though the validated labels still may not be 100\% accurate.

\subsection{Noise patterns}
\label{sec:analysis}

During the validation process, each note is assigned to two or more validating coders. We can measure inter-coder agreement rate between the validating coders as well as the agreement rate between the original coders and the validating coders. 
Given two coders and the set of the notes that are annotated by the both coders, the agreement rate is the rate of the notes that two coders annotated exactly the same code set.
The average agreement rate between the original and validating coders is 65\%, and the average agreement rate between the two validating coders is 72\%. Note that validating coders can see the original annotations, which may have contributed to the higher agreement rate. 

\newcommand{\oricoder}{the original coder}
\newcommand{\validator}{the validating coder}

Second, we categorize the agreement and disagreement between the original code set $S_o$ and the validated code set $S_v$ as follows:
\begin{itemize}
    \item match: $S_o=S_v$
    \item missing: $S_o\subset S_v$
    \item extra: $S_o \supset S_v$
    \item replacement: $|S_o - S_v|=1$, $|S_v - S_o|=1$ (one code in $S_o$ is replaced by another code in $S_v$)
    \item other: all other cases 
\end{itemize}

Table~\ref{tab:code_set_category} shows an example for each category as well as the percentage of each category in our dataset. We can see that the most common form of disagreement is ``replacement''. So we focus on notes from this category in which one code is replaced with another, and we want to see how similar or confusing the code pairs are. 
We leverage the ICD-10 code hierarchy, and check how each pair of codes differ in the hierarchy by computing the length of the common prefixes in the code. 

The top level of the hierarchy, indicated by the first letter, is referred to as \textit{chapter}\footnote{There are a few chapters whose code cover two alphabet characters (e.g., C and D are within the same chapter.) In this case, we consider these two characters as the same character.}. The first three characters together are referred to as \textit{category}. Each additional character adds more specificity to the code definition.
Table~\ref{tab:common_prefix} shows the breakdown results based on the common characters for the ``replacement'' type of disagreement. The code pairs under completely different chapters (first character) or different categories (three characters) contribute to 67\% of the disagreement, and the disagreement on minor details is only 33\%. This is somewhat surprising as one might expect coders tend to disagree with the subtle differences at the lower levels of the hierarchy (characters four, five or six). Further investigations show that among the 40.9\% of the cases that have different chapters, 19.4\% involve the ``R'' chapter in one of the codes. The ``R'' chapter stands for ``symptoms, signs, and abnormal clinical and laboratory findings, not elsewhere classified''~\cite{icdguideline}.

\begin{table*}[htbp]
\centering
\begin{tabular}{c|c|c|c}
\toprule 
\textbf{Common Characters} & \textbf{Difference}        & \textbf{Ratio} & \textbf{Example}            \\ \hline
0                 & chapter           & 40.9\% & \ul{S83.241}  $\rightarrow{}$ \ul{J80}        \\
1                 & category          & 23.7\% & M\ul{54.6} $\rightarrow{}$  M\ul{47.816}         \\
3                 & subcategory       & 17.5\% & M54.\ul{5} $\rightarrow{}$  M54.\ul{6}     \\
4                 & detail            & 16.1\% & M47.8\ul{2} $\rightarrow{}$  M47.8\ul{16}    \\
5 or 6              & detail & 1.9\%  & M47.81\ul{7} $\rightarrow{}$  M47.81\ul{6} \\ \bottomrule
\end{tabular}
\caption{The breakdown of the confused code pairs based on how characters they share. }
\label{tab:common_prefix}
\end{table*}

\section{Experiment Setup}
\label{sec:howdoeslabelnoise}
We split the data with validated labels into development (dev) and test set and use the majority of the data without validated labels as training set which resulting in 174,532 instances within the training set and 1,688 instances within the dev and test sets.



We extracted 128,771 bag of n-gram features from the notes, and trained binary logistic regression classifiers for top 100 most common ICD-10 code separately using cross-entry loss, in a one-vs-all setup. Note that there exist methods that leverage label dependencies during training as opposed to treating them independently, but we did not find that to be beneficial in our dataset and did not explore that direction. 


For evaluation, we compute average precision (AP) for each code, which measures the quality of the ranked list produced by the classifier, and is well suited for tasks with imbalanced label distributions. We then average over all the codes and report mean average precision (MAP). Because for each test instance we have both the original labels and the validated labels, we compute two MAP scores treating each of them as ground truth, in order to understand how labeling noise affects classifier evaluation. Note that classifier training is always performed based on the original labels. 

\subsection{Observations}

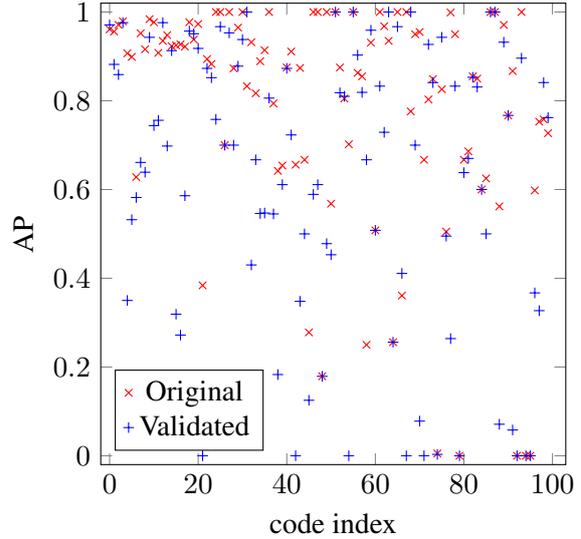
\begin{figure}[ht]
\begin{tikzpicture} 
\begin{axis}[width=\columnwidth, height=\columnwidth,
xmax=103, xmin=-2,
ymax=1.02, ymin=-0.02,
xlabel={code index},
ylabel={AP},
legend pos=south west,
cycle list name=color list
]
\addplot[only marks, mark=x, color=red]
coordinates {
(0, 0.961)
(1, 0.956)
(2, 0.971)
(3, 0.979)
(4, 0.907)
(5, 0.899)
(6, 0.628)
(7, 0.952)
(8, 0.916)
(9, 0.984)
(10, 0.977)
(11, 0.908)
(12, 0.935)
(13, 0.948)
(14, 0.923)
(15, 0.925)
(16, 0.927)
(17, 0.922)
(18, 0.977)
(19, 0.938)
(20, 0.973)
(21, 0.384)
(22, 0.894)
(23, 0.883)
(24, 1)
(25, 1)
(26, 0.7)
(27, 1)
(28, 0.873)
(29, 0.965)
(30, 1)
(31, 0.833)
(32, 0.932)
(33, 0.817)
(34, 0.889)
(35, 0.914)
(36, 1)
(37, 0.794)
(38, 0.642)
(39, 0.654)
(40, 0.875)
(41, 0.911)
(42, 0.656)
(43, 0.874)
(44, 0.667)
(45, 0.278)
(46, 1)
(47, 1)
(48, 0.179)
(49, 1)
(50, 0.568)
(51, 1)
(52, 0.875)
(53, 0.805)
(54, 0.702)
(55, 1)
(56, 0.863)
(57, 0.855)
(58, 0.25)
(59, 0.931)
(60, 0.508)
(61, 1)
(62, 0.968)
(63, 0.935)
(64, 0.256)
(65, 1)
(66, 0.361)
(67, 1)
(68, 0.776)
(69, 0.95)
(70, 0.955)
(71, 0.667)
(72, 0.803)
(73, 0.849)
(74, 0.005)
(75, 0.826)
(76, 0.505)
(77, 0.999)
(78, 0.95)
(79, 0)
(80, 0.667)
(81, 0.686)
(82, 0.855)
(83, 0.85)
(84, 0.6)
(85, 0.625)
(86, 1)
(87, 1)
(88, 0.562)
(89, 0.971)
(90, 0.767)
(91, 0.867)
(92, 0)
(93, 1)
(94, 0)
(95, 0)
(96, 0.598)
(97, 0.753)
(98, 0.758)
(99, 0.727)
};
\addlegendentry{Original}
\addplot[only marks, mark=+, color=blue]
coordinates {
(0, 0.971)
(1, 0.882)
(2, 0.859)
(3, 0.976)
(4, 0.35)
(5, 0.532)
(6, 0.582)
(7, 0.661)
(8, 0.639)
(9, 0.943)
(10, 0.744)
(11, 0.756)
(12, 0.976)
(13, 0.698)
(14, 0.913)
(15, 0.319)
(16, 0.272)
(17, 0.586)
(18, 0.957)
(19, 0.951)
(20, 0.918)
(21, 0)
(22, 0.873)
(23, 0.852)
(24, 0.758)
(25, 0.967)
(26, 0.7)
(27, 0.953)
(28, 0.7)
(29, 0.878)
(30, 0.938)
(31, 1)
(32, 0.43)
(33, 0.667)
(34, 0.546)
(35, 0.547)
(36, 0.806)
(37, 0.545)
(38, 0.183)
(39, 0.611)
(40, 0.873)
(41, 0.723)
(42, 0)
(43, 0.348)
(44, 0.5)
(45, 0.125)
(46, 0.589)
(47, 0.611)
(48, 0.179)
(49, 0.478)
(50, 0.453)
(51, 1)
(52, 0.818)
(53, 0.81)
(54, 0)
(55, 1)
(56, 0.903)
(57, 0.819)
(58, 0.667)
(59, 0.959)
(60, 0.508)
(61, 0.833)
(62, 0.729)
(63, 1)
(64, 0.256)
(65, 0.967)
(66, 0.411)
(67, 0)
(68, 1)
(69, 0.7)
(70, 0.078)
(71, 0)
(72, 0.927)
(73, 0.841)
(74, 0.003)
(75, 0.943)
(76, 0.495)
(77, 0.264)
(78, 0.833)
(79, 0)
(80, 0.638)
(81, 0.67)
(82, 0.853)
(83, 0.831)
(84, 0.6)
(85, 0.5)
(86, 1)
(87, 1)
(88, 0.071)
(89, 0.932)
(90, 0.767)
(91, 0.058)
(92, 0)
(93, 0.896)
(94, 0)
(95, 0)
(96, 0.367)
(97, 0.327)
(98, 0.841)
(99, 0.762)
};
\addlegendentry{Validated}
\end{axis}
\end{tikzpicture} 
\caption{The model's AP over codes on original labels and validated labels. The APs on validated labels (blue plus sign) often appears in the lower part of the graph, while the APs on original labels (red cross sign) mostly appears in upper part of the graph. The experiment was done on development set. }
\label{fig:cdh_full}
\end{figure}

MAP of top 100 ICD-10 codes is measured on the development set.
The MAP of the classifiers evaluated against the original labels is 88.5\%. The MAP decreases to 48.1\% when evaluated against the validated labels. Figure~\ref{fig:cdh_full} shows the model's average precision on each code evaluated on the original and validated labels. For several ICD-10 codes, scores on the validated labels are  much lower than scores on the original labels. Table~\ref{tab:summary} summarizes the score changes on all ICD-10 codes. While the overall MAP change is about 40\%, only 32 ICD-10 codes show significant changes that are larger than 20\%.

This indicates that score changes are not uniform across all ICD-10 codes, and rather it is concentrated on a few codes.

Previous studies show that if the label noise is random and independent of features or labels, classifier's performance evaluated on cleaned labels should be higher than it's performance evaluated on noisy labels ~\cite{forward,ma2018dimensionality,sce}. Here we observe the opposite trend. This suggests that the noise in our data may not be random and independent, but systematic. 

We inspected the cases on which validating coders disagree with the original coders and observed that there exists systematic noise in the original labels and is fixed by validating coders. One prevalent noisy pattern is that the original coders often assign symptom ICD-10 codes when diagnosis ICD-10 codes should have been assigned.
In ICD-10, there are codes that are used to indicate specific \textit{diagnoses} (e.g., effusion on knee, spinal stenosis or intervertebral disc displacement), as well as codes that indicate \textit{signs or symptoms} (e.g., fever, pain in back or cough). The official ICD-10 guideline says that 
``Codes that describe symptoms and signs, as opposed to diagnoses, are acceptable for reporting purposes when a related definitive diagnosis has {\bf not} been established (confirmed) by the provider.''~\cite{icdguideline}.

\begin{table}[htbp]
\centering
\begin{tabular}{c|c}
\toprule
 \textbf{Evaluated Label} & \textbf{MAP} \\ \hline
 original (O)                          & 0.885 \\ \hline 
 validated (V)                        & 0.481 \\ \hline 
 \multicolumn{2}{c}{} \\ \hline 
\textbf{Score Change ($\Delta$)} & \textbf{\# Codes} \\ \hline 
$\Delta > 0.2$               & 32      \\ \hline
 $0.2 \geq \Delta > 0.05 $  & 18      \\ \hline
 $0.05 \geq \Delta > -0.05 $  & 42      \\ \hline
 $-0.05 \geq \Delta $                 & 8      \\ \hline
 total & 100 \\ \bottomrule
\end{tabular}
\caption{Summary of the baseline model's performance in original and validated labels. The score differences were larger in some of the codes, where in many codes have little or no difference. The experiment was done on development set. }
\label{tab:summary}
\end{table}



Among the 32 ICD-10 codes that have AP score changes above 20\%, 16 codes are symptom codes. When the classifier is trained on labels with such systematic noise, it learns to also over-predict symptom codes on cases that should have been assigned diagnosis codes, and therefore, receives low AP when evaluated on validated codes. For example, table~\ref{tab:m545_top} shows the top 10 instances ranked by classifier's prediction confidence for the code \textit{M54.5} (``Low back pain''). Many instances with false positive \textit{M54.5} predictions have in their ground truth specific diagnosis codes that are related to the \textit{M54.5} symptom code. Table~\ref{tab:m545_rel_desc} shows the descriptions for the related diagnosis codes. 

\begin{table}[htbp]
\resizebox{\columnwidth}{!}{
\begin{tabular}{c|c|c|l}
\toprule 
\textbf{Rank} & \textbf{Doc ID} & \textbf{Correct}      & \multicolumn{1}{c}{\textbf{Validated Labels}}               \\ \hline
1    & 1250   & 1 & \textbf{M54.5}, M47.817           \\
2    & 497    & 0          & M47.817                  \\
3    & 210    & 0          & M51.26, M47.816, M48.062 \\
4    & 1579   & 0          & M48.061, M51.16          \\
5    & 584    & 0          & M51.26, M48.061, M43.17  \\
6    & 417    & 1 & \textbf{M54.5}                    \\
7    & 996    & 0          & M48.07, N63.12           \\
8    & 1210   & 0          & M41.86                   \\
9    & 1423   & 1 & \textbf{M54.5}, M51.26            \\
10   & 294    & 0          & M48.061              \\
\bottomrule
\end{tabular}
}
\caption{Top ranked instances for the code M54.5. The column `correct' indicates whether the validated labels contain M54.5}
\label{tab:m545_top}
\end{table}

While diagnosis-symptom confusion is most common in this dataset, there are other types of code pairs that may cause confusion. In general, we refer to the code we are trying to predict as the \textit{target code} and the code that cause confusion as the \textit{confusion code}. In the noisy dataset, some instances are wrongly annotated with the target codes while their clean labels include the \textit{confusion codes}. This causes the model to incorrectly predict the target code on instances where confusion codes should have been assigned. In the previous example, the symptom codes are the target codes and the diagnosis codes are the confusion codes. 

\begin{table*}[htbp]
\centering
\begin{tabular}{l|l}
\toprule
 \textbf{Code}      & \textbf{Description}                                                                    \\ \hline
M54.5            & Low back pain                                                                  \\\hline \hline
M41.86           & Other forms of scoliosis, lumbar region                                       \\ 
M47.816          & Spondylosis without myelopathy or radiculopathy, lumbar region\footnotemark[1]                 \\
M47.817          & Spondylosis without myelopathy or radiculopathy, lumbosacral region\footnotemark[2]            \\
M48.061          & Spinal stenosis, lumbar region without neurogenic claudication                 \\
M51.26           & Other intervertebral disc displacement, lumbar region                          \\
M51.16           & Intervertebral disc disorders with radiculopathy, lumbar region                \\ \bottomrule
\end{tabular}
\caption{ICD code descriptions for M54.5 and codes for specific diagnosis that are related to M54.5.\\
\footnotesize{\footnotemark[1]lumbar: relating to the lower part of the back\\
\footnotemark[2]lumbosacral: relating to the lumbar and sacral regions or parts
}
}
\label{tab:m545_rel_desc}
\end{table*}

The opposite scenario where the specific diagnosis is the target code and the symptom codes are confusion codes is also possible but not as common, and therefore, we do not handle such scenario in this paper.


\section{Handling Label Noise in Training}
\label{sec:solution}
Our proposed training method first identifies which codes are often confused with our \textit{target codes} based on the validated codes of the development split, and then supervised those codes (\textit{confusion codes}) along with the target codes so that the model is able to better distinguish them. When the codes for generic symptoms are the target codes, the codes for specific diagnoses work as confusion codes. 



\subsection{Selecting confusion codes}
For each of the target codes that are identified to have systematic noise, we select confusion codes based on the code descriptions and the model's behavior on the development set. 

First, for each target code, we train an initial binary classifier on original labels as described in the previous section, and we apply it to each instance in development split to generate a label probability, based on which can rank all instances w.r.t. this target code.
From the top-k instances, we further select those instances on which the classifier generates false positive predictions w.r.t. the target code. We used $k$ value of 50. Next, we check which other codes are assigned to these selected instances by validating coders and consider them as \textit{candidate codes}. 

Let $y^{(i)}$ be the code set for $i^{th}$ ranked instance. Let $p_i$ be the model predicted probability of $i^{th}$ instance containing the target code $t$. Candidate code set $C$ is given as 
\begin{align}
C = \{c | c \in y^{(i)}, t \not\in y^{(i)}, 0.5 \leq p_i, i \leq k \}. 
\end{align}

Then, each candidate code $c$ is scored by the sum of the reciprocal rank of the instances that is annotated with candidate code. 
\begin{align}
    S_c = \sum_i^k \frac{1}{i} \cdot \mathbbm{1}[c \in y^{(i)} \text{and } t \notin y^{(i)} ],
\end{align}
where $\mathbbm{1}$ is the indicator function. 
We then keep the codes whose scores are larger than 0.1.

Next, the selected candidate codes are filtered by their locations in the hierarchy, based on two simple heuristics: 1) accept code if the candidate code is from the same chapter as the target code; 2) accept code if either the candidate code or the target code is from chapter ``R'' or ``S''. Our intuition is that the codes from the same chapters are more likely to be confusing. 
The ``R'' chapter (symptoms, signs and abnormal clinical and laboratory findings, not elsewhere classified) and the ``S'' chapter (injury, poisoning and certain other consequences of external causes) could act as \textit{wastebasket} chapters in the Radiology domain and coding professionals often struggle with these codes~\cite{butler2016analyzing}. 
This design agrees with the observation in section~\ref{sec:analysis} that the most confusing pairs appear in the same or limited chapters. Even though these two heuristics are very simple, we observe them to work well when combined with the first score-based filtering. 

\subsection{Modified supervision}
As noted in previous sections, the classifier training data contains only the noisy labels. As a result, we explicitly model the label noise during model training. We map each of the training instances to one of the following three classes based on its original labels.
\begin{itemize}
    \item class 0: None of the target code or confusion codes is in original labels.
    \item class 1: The target code is in original labels, and no confusion code is in original labels.
    \item class 2: At least one confusion code is in original labels.
\end{itemize}
We separate class 2 from class 1 because instances in class 2 are often mislabeled with the target code.  
We then train a standard 3-class logistic regression classifier using the mapped classes. During prediction, we use the prediction score for class 1 to determine the target code and ignore scores from class 0 and class 2.

\section{Experiments}
\newcommand{\modfiedlabelmethod}{modified label}
\newcommand{\Modfiedlabelmethod}{Modified label}

We evaluated the proposed training method on the 16 of ICD-10 codes which we identified to have label noise problems (section~\ref{sec:analyzing}). 
\subsection{Baselines}
We compare our method with the following baselines:
\begin{itemize}
    \item Vanilla: standard logistic regression trained on the original noisy labels
    \item \Modfiedlabelmethod: if a training instance has both target code and confusion codes in the original labels, only keep the confusion codes
    \item Dev set fine-tuned: first train the classifier on training set using original labels, and then fine tune on development set using validated labels
    \item Dev set trained: train a classifier on the small development set using the validated codes 
    \item MLP + BR: train a multi-layer perceptron with one hidden layer and $(n+1)$-output nodes for one target code and $n$ confusion code. This model does not assume exclusive relations between codes and is expected to capture certain code dependencies as the hidden layer is shared.
\end{itemize}

\subsection{Results}
A summary of different classifiers' performance on the test set is given in Table~\ref{tab:main_exp} using two versions of labels: the original codes and the validated codes.

\begin{table}[htbp]
\centering
\begin{tabular}{l|c|c}
\toprule
\textbf{Method}             & \textbf{Original} & \textbf{Validated} \\ \hline
Vanilla           & \textbf{88.5}    & 48.1     \\ \hline
\multicolumn{3}{c}{w/ confusion codes}   \\ \hline
\Modfiedlabelmethod       & 72.9    & 60.0     \\ 
Proposed           & 73.5    & \textbf{63.3}     \\
MLP + BR           & 79.4    & 57.8     \\ \hline
\multicolumn{3}{c}{w/o confusion codes}  \\ \hline
Dev set trained    & 9.5    & 8.0     \\ 
Dev set fine-tuned & 76.9    & 53.8    \\
\bottomrule
\end{tabular}
\caption{MAP for all the methods. The ultimate goal is to get a high score on validated labels. The score on original labels are listed only for reference.  Statistical significance for a few pairs are: 9.9\% between \textit{vanilla} and \textit{proposed} on validated, 12.4\% between \textit{\modfiedlabelmethod} and \textit{proposed} on validated and  39.2\% between \textit{\modfiedlabelmethod} and \textit{proposed} on original.
}
\label{tab:main_exp}
\end{table}

The methods' performance on validated labels and the original labels are negatively correlated, matching our expectation that the noise is systematic -- classifiers mimicking human mistakes will match original labels well but not validated labels. When evaluated against validated codes, our proposed method outperforms the rest.

The results also show that the methods which explicitly consider confusion codes (\textit{\modfiedlabelmethod}, \textit{proposed} and \textit{MLP+BR}) perform better than those who do not. Furthermore, the \textit{proposed} method, which trains the model with 3-way categorical cross-entropy shows better score over using 2-way cross-entropy loss (\textit{\modfiedlabelmethod}). This shows that our method does more than fixing training labels using a simple rule. \textit{MLP+BR} is better than the \textit{vanilla} but worse than our model. This shows that simply feeding relevant label information to a neural network without imposing a strong inductive bias based on the noise pattern does not guarantee successful training. 

While not listed in table, we have also tried a noise-handling approach such as symmetric cross-entropy~\cite{sce}, which showed worse score (63.3) than the \textit{vanilla} method (64.8)\footnote{This experiment was done on the different data split}.

Finally, \textit{dev set trained} performs poorly due to the limited training data. \textit{dev set fine-tuned} outperforms \textit{vanilla} but cannot match the methods that consider confusion codes.

\section{Discussion}
Several independent studies have reported that the  inter-coder agreement for clinical billing is low.
American Health Information Management Association (AHIMA) reported that the regular trained coders have about 42\% accuracy evaluated against experts' annotations~\cite{american2003icd}. In a study on ICD coding for ambulatory visits, it is reported only half of entered codes were actually appropriate and about a quarter codes were missing~\cite{horsky2017accuracy}.

The most commonly used datasets for ICD classification are  MIMIC-III~\cite{johnson2016mimic} and CLEF~\cite{neveol2017clef}. To the best of our knowledge, no subsequent label quality check or cleaning has been performed on these public datasets, and therefore, we did not use these datasets for studying label noise.

Previous studies on ICD code classification tasks mainly focused on exploring different machine learning models for the task. A number of neural methods, including LSTM~\cite{ZENG201943}, attention mechanism~\cite{falis2019ontological}, and pre-trained language models~\cite{amin2019mlt} have been proposed for such task. This work concentrates on an orthogonal aspect of ICD classification -- label noise. Our modified training strategy is a general method and can be combined with different types of classifiers. 

There are many existing works on training noise-robust classifiers~\cite{bucak2011multi, natarajan2013learning, liu2015classification, izadinia2015deep, zhao2015semi, forward}. Some methods make over-simplified assumptions about the noise and are not suitable for our problem (see  section~\ref{sec:analysis}). For example, several methods assume the noise is completely random and is independent of features or labels~\cite{sce}; and several others assume the noise is only in the form of missing labels~\cite{kanehira2016multi, jain2017scalable}.  There are also methods that leverage samples with both noisy labels and clean labels in order to estimate the noise pattern~\cite{xiao2015learning, vahdat2017toward, veit2017learning}. Unlike previous works, our design of the noise handling method is driven both by the data and clinical domain knowledge (i.e., ICD hierarchy). 


\section{Conclusion and Future Work}
This paper investigates the characteristics of labeling noise in manually-assigned ICD-10 codes and furthermore, proposes a method to train robust ICD-10 classifiers in the presence of labeling noise. Our research concluded that the nature of such noise is systematic. We compared our method with several different baselines including one that does not handle label noise and the baseline methods that assume random noise, and demonstrated that our proposed method outperforms all baselines when evaluated on expert validated labels.

We are currently working on extending our proposed method for data from other service lines besides Radiology, such as Pathology. We are also investigating the impact of including coders from different hospitals that may follow different coding conventions and as a result, may code the same note differently. Explicit modeling of such inter-site disagreement during model training will help us generate a more robust and scalable model. 


\bibliography{main}
\bibliographystyle{conf_paper/acl_natbib}

\end{document}